\pdfoutput=1

\documentclass[11pt]{article}

\usepackage[final]{acl}

\usepackage{times}
\usepackage{latexsym}

\usepackage[T1]{fontenc}

\usepackage[utf8]{inputenc}

\usepackage{microtype}

\usepackage{inconsolata}

\usepackage{graphicx}

\usepackage{amsmath}
\usepackage{graphicx}
\usepackage{booktabs}
\usepackage{transparent}
\usepackage{multirow}
\usepackage[capitalize]{cleveref}
\usepackage[super]{nth}
\usepackage[shortlabels]{enumitem}
\usepackage{quoting}
\usepackage{subcaption}
\usepackage{adjustbox}
\usepackage{xcolor}
\usepackage{ulem}
\usepackage{scalerel}
\usepackage{dcolumn}
\usepackage{url}
\usepackage{array}
\usepackage{makecell}
\usepackage[super]{nth}

\usepackage{stmaryrd}

\usepackage{amssymb}
\usepackage{pifont}

\definecolor{green}{RGB}{78,167,46}
\definecolor{red}{RGB}{192,0,0}
\definecolor{yellow}{RGB}{255,192,0}

\newcolumntype{d}[1]{D{.}{.}{#1}}
\newcommand{\green}[1]{\textcolor{green}{#1}}
\newcommand{\red}[1]{\textcolor{red}{#1}}
\newcommand{\warning}[1]{\textcolor{red}{#1}}

\crefformat{section}{\S#2#1#3} 
\crefformat{subsection}{\S#2#1#3}
\crefformat{subsubsection}{\S#2#1#3}

\newcommand{\quotes}[1]{``#1''}

\newcommand{\passage}{\texttt{\{Text\}}}
\newcommand{\keyword}{\texttt{\{Keyword\}}}
\newcommand{\gloss}{\texttt{\{Gloss\}}}

\title{Bias Analysis and Mitigation through Protected Attribute Detection and Regard Classification}

\author{Takuma Udagawa$^1$, Yang Zhao$^1$, Hiroshi Kanayama$^1$,  Bishwaranjan Bhattacharjee$^2$ \\
$^1$IBM Research - Tokyo \; \; $^2$IBM T. J. Watson Research Center \\
\texttt{\{takuma.udagawa,yangzhao\}@ibm.com \; hkana@jp.ibm.com \; bhatta@us.ibm.com}}

\begin{document}
\maketitle
\begin{abstract}
Large language models (LLMs) acquire general linguistic knowledge from massive-scale pretraining. 
However, pretraining data mainly comprised of web-crawled texts contain undesirable social biases which can be perpetuated or even amplified by LLMs.
In this study, we propose an efficient yet effective annotation pipeline to investigate social biases in the pretraining corpora.
Our pipeline consists of \textit{protected attribute detection} to identify diverse demographics, followed by \textit{regard classification} to analyze the language polarity towards each attribute.
Through our experiments, we demonstrate the effect of our bias analysis and mitigation measures, focusing on Common Crawl as the most representative pretraining corpus.

\warning{Warning: This paper contains examples of social biases that may be harmful or offensive.}

\end{abstract}

\section{Introduction}
\label{sec:introduction}

\begin{figure*}[t!]
    \includegraphics[width=0.99\textwidth]{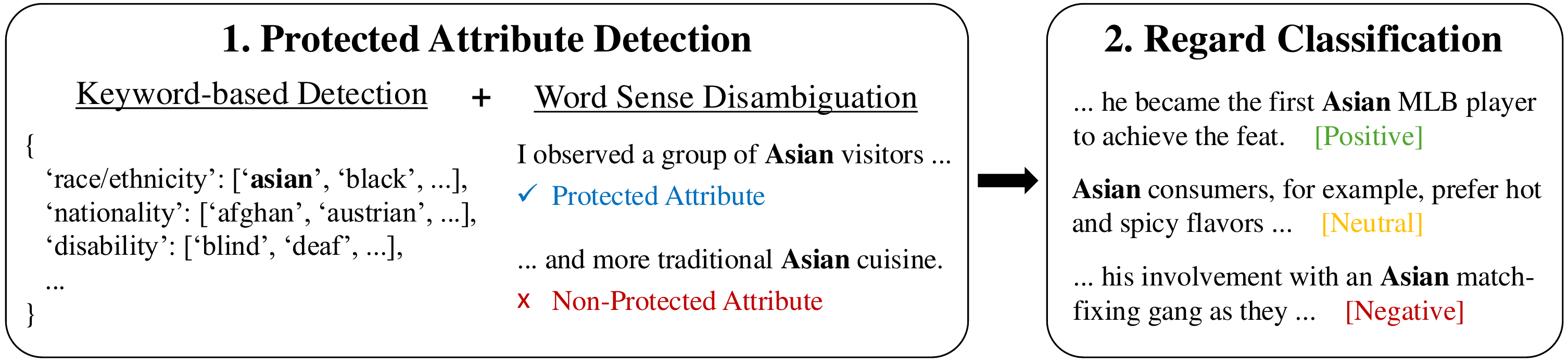}
    \caption{
    An overview of our annotation pipeline consisting of (1) \textit{protected attribute detection} to identify diverse demographics and (2) \textit{regard classification} to analyze the language polarity towards each detected attribute.
    }
    \label{fig:overall_pipeline}
\end{figure*}

Recent years have witnessed a remarkable progress in the development and adoption of large language model (LLM) technology \citep{achiam2023gpt,dubey2024llama}.
Generally, LLMs first undergo large-scale pretraining to acquire general linguistic knowledge, followed by post-training to align them with human goals and preferences \citep{ouyang2022training}.
The majority of LLM's knowledge is acquired in the pretraining stage, and post-training is conceived to mainly change the style of LLMs to serve as interactive, open-domain AI assistants \citep{zhou2024lima,lin2024the}.

However, pretraining data comprised of web-crawled texts often contain undesirable biases, such as associating Muslim people with terrorism and extremism \citep{chowdhery2023palm}.
Consequently, LLMs tend to inherit the biases and produce harmful judgements \cite{sheng-etal-2021-societal,schramowski2022large}, raising significant risks in terms of safety and fairness.
Despite the severity of this issue, it remains extremely difficult to audit and alleviate such dataset-level biases due to the ever-increasing size of pretraining corpora and the open-ended, complex nature of social biases. 

In this study, we make a first step towards analyzing and mitigating problematic social biases in massive-scale text data.
Specifically, we propose a scalable annotation pipeline consisting of two steps. 
First, we conduct \textit{protected attribute detection} to identify diverse demographics, such as nationality, religion, disability, etc.
To reduce false positive detections, we combine keyword matching and word sense disambiguation \citep{huang-etal-2019-glossbert}.
Second, we apply \textit{regard classification} to analyze the language polarity towards each attribute into positive, neutral, or negative \citep{sheng-etal-2019-woman}.
Our overall pipeline is illustrated in Figure \ref{fig:overall_pipeline}.

In our experiments, we apply the pipeline to a subset of Common Crawl, the most widely used corpus for LLM pretraining.
For bias analysis, we refine the frequency-based word association analysis \citep{bordia-bowman-2019-identifying} to take into account regard information.
Qualitative and quantitative results show that our approach more accurately captures commonly held stereotypes \citep{jha-etal-2023-seegull}. 
For bias mitigation, we reveal that the regard distributions among protected attributes can be noticeably imbalanced, and adjusting regard distributions (e.g.\ downsampling negative regard texts) can be a promising approach to mitigate offensive social stereotypes, such as \quotes{\textit{white}} people being overly described as \quotes{\textit{racists}} or \quotes{\textit{supremacists}}.

While several challenges still remain, we expect our approach to be a crucial step towards improving the fairness of LLM pretraining datasets.

\begin{table*}[t]
\centering
\begin{adjustbox}{max width=0.99\textwidth}
\begin{tabular}{lc}
\hline
\multicolumn{1}{c}{Input Format: [BOS] \passage [SEP] \keyword; \gloss [EOS]} & Prediction \\
\hline
\hline
[BOS] I observed a group of \textbf{Asian} visitors ... [SEP] \textbf{asian}; a person of asian race/ethnicity [EOS] & \checkmark \,Protected Attribute \\
\hline
[BOS] ... and more traditional \textbf{Asian} cuisine. [SEP] \textbf{asian}; a person of asian race/ethnicity [EOS] & $\times$ \,Non-Protected Attribute \\
\hline
\end{tabular}
\end{adjustbox}
\caption{Word sense disambiguation in the style of Gloss-BERT. Regarding the input format, \passage \, indicates the text containing the keyword, \keyword \, the target keyword, and \gloss \, the keyword's defined usage (gloss).}
\label{tab:gloss_roberta}
\end{table*}

\section{Methods}
\label{sec:methods}

\subsection{Protected Attribute Detection}
\label{subsec:protected_attribute_detection}

To efficiently detect protected attributes in massive pretraining data, existing work typically relies on keyword matching \citep{chowdhery2023palm,esiobu-etal-2023-robbie}.
Following this line, we defined a taxonomy of protected attributes covering 10 diverse classes containing a total of 97 keywords.
For instance, \quotes{\textit{asian}} is a keyword belonging to the class of \textit{race/ethnicity} (cf. Figure \ref{fig:overall_pipeline}).
We include more details of our taxonomy in Appendix \ref{subsec:taxonomy}.

However, keywords representing protected attributes are often polysemous and yield many false positives. 
For instance, \quotes{\textit{blind}} can indicate a person's disability (visual impairment) or used in a totally different sense (e.g.\ \textit{blind date}).
For reliable bias measurement, we must accurately recognize and compare the relevant \textit{human} attributes.\footnote{\citet{blodgett-etal-2021-stereotyping} also emphasize the importance of comparing the commesurables, e.g.\ avoiding conflation of stereotypes about Norwegian \textit{people} and \textit{salmon}.}

To efficiently reduce false positives, we conduct word sense disambiguation (WSD) in the style of Gloss-BERT \citep{huang-etal-2019-glossbert}.
In this approach, WSD is framed as a binary classification of whether a keyword in a text is used following the sense described in a gloss.
We adopt this framework to predict whether a keyword is used to indicate a defined protected human attribute (Table \ref{tab:gloss_roberta}).

While traditional WSD relies on glosses defined in WordNet \cite{miller1995wordnet}, this is not suitable for our purpose, e.g.\ since there are no sense inventories which distinguish the usages of \quotes{\textit{asian}} in Table \ref{tab:gloss_roberta}.
Therefore, for each of the 97 keywords, we crafted simple yet sufficient glosses describing the protected attributes.\footnote{It is worth noting that our glosses are NOT intended to define strict word senses but to practically distinguish whether the keyword indicates a protected \textit{human} attribute.}
We show several illustrative examples in Table \ref{tab:taxonomy} (Appendix \ref{subsec:taxonomy}). 

To develop our WSD model, we used an existing LLM to synthesize high-quality training dataset.
Specifically, for each keyword, we randomly extracted 1K sentences from Common Crawl which contain the keyword.
Then, we leveraged Mixtral-8x22B-Instruct\footnote{\url{https://huggingface.co/mistralai/Mixtral-8x22B-Instruct-v0.1}} \citep{jiang2024mixtral} to collect judgements on whether the keyword indicates the defined protected attribute using the prompt in Table \ref{tab:mixtral_prompts}.
Based on this dataset, we trained a light-weight WSD model from RoBERTa$_{\scriptstyle \,\textsc{BASE}}$ \citep{liu2019roberta} for scalable inference, which we refer to as Gloss-RoBERTa in the rest of this paper.

\begin{table*}[t!]
\setlength\extrarowheight{-1.5pt}
\centering
\begin{adjustbox}{max width=1.0\textwidth}
\setlength\tabcolsep{3.5pt}
\begin{tabular}{
    cllll
}
\hline
\multirow{2}{*}[-0.7ex]{\shortstack[c]{Protected\\Attribute}} & \multicolumn{1}{c}{\multirow{2}{*}[-0.7ex]{\shortstack[c]{Frequency Bias\\(eq. (\ref{eq:bias_score}))}}} & \multicolumn{3}{c}{\raisebox{-0.2ex}{Frequency+Regard Bias (eq. (\ref{eq:regard_bias_score}))}} \\
\cmidrule(lr){3-5}
 & & \multicolumn{1}{c}{$r=$ Positive} & \multicolumn{1}{c}{$r=$ Negative} & \multicolumn{1}{c}{$r=$ Neutral} \\
\hline
\hline
\multirow{3}{*}{arab} & {\small arab, palestinian, israeli, syrian,} & {\small idol, mars, astronaut, blasted, gener-} & {\small terrorist, assaults, wounded, bomb,} & {\small denomination, consult,} \\
& {\small israel, iraqi, arabs, lebanon, iraq,} & {\small osity, chairperson, orchestra, poets,} & {\small destroy, destruction, towers, terror,} & {\small mingle, filter, traded,} \\
& {\small lebanese, egypt, egyptian, ...} & {\small praised, forbes, hailed, hospitality, ..} & {\small attacks, civilians, destroyed, ...} & {\small tagged, assessing, ...} \\
\hline
\multirow{3}{*}{black} & {\small bipoc, starbucks, unarmed, abr-} & {\small vogue, essence, untold, uplifting, o-} & {\small fatally, cops, cop, breathe, shot,} & {\small under-represented, com-} \\
& {\small ams, brutality, freddie, custody,} & {\small prah, creatives, hidden, salute, amp-} & {\small gun, tear, misconduct, surfaced,} & {\small plexion, applications, d-} \\
& {\small black, systemic, fatally, ...} & {\small lify, pulitzer, congresswoman, ...} & {\small mentally, fired, killing, falsely, ...} & {\small isabilities, disabled, ...} \\
\hline
\multirow{3}{*}{white} & {\small supremacist, collar, privileged,} & {\small guy, teammates, blues, educated,} & {\small supremacist, racists, mob, breathe,} & {\small makeup, races, followed,} \\
& {\small blond, settlers, privilege, evan-} & {\small refusal, dude, afforded, hunters,} & {\small roof, pleaded, raped, brutally, angr-} & {\small weighs, pounds, inches,} \\
& {\small gelicals, privileges, white, ...} & {\small kid, parks, priviledge, loving, ...} & {\small ry, guilty, murdered, resentment, ...} & {\small weighing, borough, ...} \\
\hline
\end{tabular}
\end{adjustbox}
\caption{Results of our bias analyses for class $A=$ \textit{race/ethnicity}. Words $w \in V$ are sorted in descending order based on the frequency bias score (eq. (\ref{eq:bias_score})) and frequency+regard bias score (eq. (\ref{eq:regard_bias_score})).}
\label{tab:bias_analysis}
\end{table*}

\subsection{Regard Classification}
\label{subsec:regard_classification}

Regard classification is widely used to evaluate the bias of LLM generated texts \citep{sheng-etal-2019-woman,bold_2021}.
In this study, we aim to apply this technique to pretraining data analysis.

Unfortunately, existing regard classifier \citep{sheng-etal-2019-woman} is only trained on templated texts (e.g.\ \quotes{\textit{XYZ was regarded as ...}}) which makes it difficult to be applied on realistic texts with diverse syntactic structures.
Therefore, similar to our WSD (\cref{subsec:protected_attribute_detection}), we generated new training data for regard classification based on an existing LLM.

Specifically, for each keyword, we used 50K sentences from Common Crawl containing the keyword after removing false positives with Gloss-RoBERTa.
Then, we used Mixtral-8x7B-Instruct\footnote{\url{https://huggingface.co/mistralai/Mixtral-8x7B-Instruct-v0.1}} to annotate the regard towards each protected attribute using the prompt in Table \ref{tab:mixtral_prompts} (Appendix \ref{subsec:mixtral_prompts}). 
This way, we synthesized a diverse and realistic dataset for regard classification, which we used to train a light-weight regard classifier from RoBERTa$_{\scriptstyle \,\textsc{BASE}}$ for efficient inference.

\subsection{Annotation Agreement}
\label{subsec:annotation_agreement}

Regarding WSD for protected attribute detection (\cref{subsec:protected_attribute_detection}), we computed the annotation agreement between Mixtral-8x22B and Gloss-RoBERTa based on Cohen's kappa.
On average across all attributes, we observed a moderate agreement of around 0.59.
Additionally, for a small set of attributes, we conducted manual annotation and confirmed that agreement (1) among human annotators and (2) between humans and these models are also at similar levels (0.56 to 0.70).
Therefore, our models can disambiguate protected attributes at near human-like levels (see Table \ref{tab:wsd_examples} for example predictions).

As for regard classification, we prepared a high-quality test set double-checked with both Mixtral-8x7B and 8x22B for consistency.
Based on our RoBERTa-based classifier, we achieved a decent F1 performance of 0.91 on micro and 0.82 on macro average across all attributes.
This verifies that we could successfully distill the regard judgements of the strong Mixtral teacher into a much more light-weight classifier (cf.\ Table \ref{tab:regard_examples}).

\section{Experiments}
\label{sec:experiments}

In our experiments, we apply our bias analysis and mitigation measures on a subset of Common Crawl (CC), a widely used LLM pretraining corpora.

As a preparation, we first sampled over 3M English documents from CC and extracted sentences of appropriate length.\footnote{Specifically, more than 16 tokens and less than 128 tokens based on the NLTK tokenizer \cite{bird2006nltk}.}
Then, we conducted protected attribute detection (\cref{subsec:protected_attribute_detection}) and regard classification (\cref{subsec:regard_classification}) on these sentences. 
In the following experiments, we used up to 100K sentences per attribute to demonstrate the effect of our bias analysis (\cref{subsec:bias_analysis}) and mitigation (\cref{subsec:bias_mitigation}) techniques.

\subsection{Bias Analysis}
\label{subsec:bias_analysis}

First, we apply the basic word association analysis following \citet{bordia-bowman-2019-identifying}.
Specifically, let $a \in A$ denote a protected attribute in class $A$ and $w \in V$ denote a word in vocabulary $V$.
Then, we can compute the \textit{frequency bias} of $w$ against $a$ based on the following score:
\begin{equation}
\frac{p(w|a)}{\llbracket\,p(w|a)\rrbracket_{a \in A}}
\label{eq:bias_score}
\end{equation}
\noindent
Here, $p(w|a)$ denotes the probability of $w$ occurring in a sentence containing $a$, and the denominator is its average for $a \in A$.
A higher score of eq. (\ref{eq:bias_score}) indicates that $w$ is more likely to co-occur with $a$ compared to other protected attributes in $A$.

In Table \ref{tab:bias_analysis}, we show (partial) results of bias analysis for $A=$ \textit{race/ethnicity}, where words are sorted in descending order based on eq. (\ref{eq:bias_score}).
Due to the limitation of space, we provide further experimental details and results in Appendix \ref{sec:bias_analysis_details}.

Unfortunately, it is often difficult to obtain useful insights on social stereotypes from frequency bias only.
For instance, words co-occurring with \quotes{\textit{arab}} are mostly proper nouns (e.g.\ \quotes{\textit{palestinian}}) which is self-evident and irrelevant to stereotypes.
Also, we can confirm words like \quotes{\textit{supremacist}} co-occur with \quotes{\textit{white}} but cannot tell whether they are used in a negative or offensive manner.

To address this issue, we propose a novel bias score reflecting both frequency and regard information.
Specifically, let $r \in R = $ \{Positive, Negative, Neutral\} denote the regard label of $a$ in the sentence.
Then, for each $r$ we compute the following score:
\begin{equation}
\min \bigl( \frac{p(w|a)}{\llbracket\,p(w|a)\rrbracket_{a \in A}},\, \frac{p(r|w, a)}{\llbracket\,p(r|w, a)\rrbracket_{r \in R}} \bigr)
\label{eq:regard_bias_score}
\end{equation}
Here, the left term represents \textit{frequency bias} (eq. (\ref{eq:bias_score})) and the right term \textit{regard bias} of when a word $w$ and attribute $a$ co-occur.\footnote{While we ignore the difference in the scale of these bias scores, this can be easily adjusted (e.g. by introducing scaling factors) if necessary.}
For instance, if $w$ = \textit{supremacist} and $a$ = \textit{white} tend to co-occur with negative regard, the score increases for $r$ = \textit{negative}.
This way, we can analyze not only whether $w$ and $a$ co-occur, but whether they co-occur accompanying specific regard in the pretraining corpora.

Table \ref{tab:bias_analysis} includes the results of our analysis based on this frequency+regard bias score.
By leveraging regard information, we can more easily identify social stereotypes such as \quotes{\textit{generosity}} (positive) and \quotes{\textit{terrorist}} (negative) towards \quotes{\textit{arab}} people.
Also, we can verify that words like \quotes{\textit{supremacist}} and \quotes{\textit{racists}} co-occur with negative regard towards \quotes{\textit{white}} people, which can be problematic if over-represented in the pretraining data.

More quantitatively, we show that our analysis better captures commonly held stereotypes listed in SeeGULL \citep{jha-etal-2023-seegull}, which is discussed in Appendix \ref{sec:bias_analysis_details}.
Overall, our approach is a simple yet effective technique to investigate social biases toward a variety of protected attributes. 

\subsection{Bias Mitigation}
\label{subsec:bias_mitigation}

\begin{figure}[t!]
    \includegraphics[width=0.99\columnwidth]{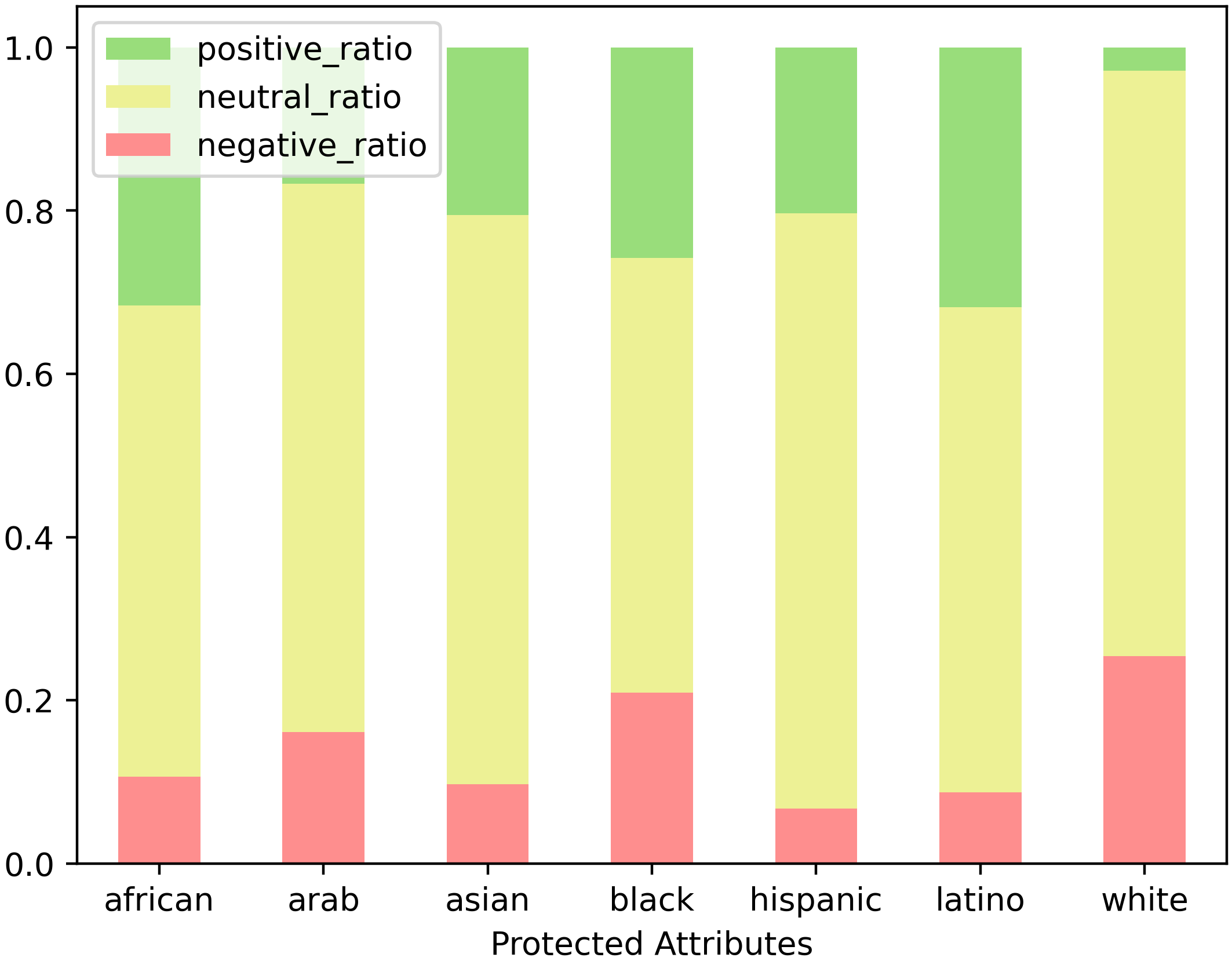}
    \caption{Regard distributions for $A=$ \textit{race/ethnicity}.
    }
    \label{fig:regard_distribution}
\end{figure}

Arguably, a fair pretraining dataset should maintain balanced regard distributions towards protected attributes.
Unfortunately, based on our regard estimation, this is not ensured in CC: for instance, \textit{white} people are much more likely to be described with negative regard in the class of $A =$ \textit{race/ethnicity}, reaching over 20\% in ratio (Figure \ref{fig:regard_distribution}).

In this study, we verify that problematic word association can be alleviated by properly balancing the regard distributions. 
Specifically, we ensure the negative regard ratio to be at most 1\% for all protected attributes by downsampling negative regard sentences.
Then, we compute the relative reduction in the conditional probability $p (w|a)$, where $w$ is a negatively biased word towards $a$.\footnote{To be precise, we report $100 \!\times\! \frac{p^\prime (w|a)}{p (w|a) }$\%, where $p$ and $p^\prime$ are probabilities before and after intervention, respectively.}

Through this intervention, we confirmed $p (w|a)$ can be dramatically reduced, e.g.\ down to 19\% and 18\% for $w =$ \quotes{\textit{supremacist}} and \quotes{\textit{racists}} toward $a =$ \quotes{\textit{white}}, respectively.
Similarly, $p (w|a)$ can be diminished to 46\% and 26\% for $w = $ \quotes{\textit{terrorist}} and \quotes{\textit{assaults}} toward $a =$ \quotes{\textit{arab}}, while neutral or positive word associations remain unchanged or slightly increase, e.g.\ up to 112\% for $w = $ \quotes{\textit{generosity}} and \quotes{\textit{hospitality}}.

It is worth noting that our approach can focus on problematic rather than benign word associations owing to the regard annotation $r$.
For instance, the former sentence will be retained while the latter is subject to downsampling, although \quotes{\textit{white}} and \quotes{\textit{racist}} co-occur in both sentences:
\begin{itemize}[topsep=0pt, itemsep=0pt, leftmargin=.2in, parsep=0pt]
  \item \textit{Not every \underline{white} person is \underline{racist} and not every black person is a criminal.} ($r =$ neutral)
  \item \textit{Report says he was attacked by three \underline{white} males at about 3:25 a.m., who made \underline{racist} comments as they assaulted him.} ($r =$ negative)
\end{itemize}

Overall, we expect our approach of regard distribution balancing to be a promising technique for mitigating undesirable social biases. 

\section{Discussion and Conclusion}
\label{sec:discussion_and_conclusion}

In this study, we proposed a novel pipeline to probe social biases in large-scale pretraining corpora.
In contrast to existing methods, our approach employs WSD to accurately detect protected attributes while maintaining efficiency.
Furthermore, we apply regard classification to analyze the language polarity towards each attribute, which is crucial yet overlooked in the existing study of pretraining datasets \citep{dodge2021documenting,penedo2024fineweb}.

Based on our annotation, we can conduct regard-aware bias analysis to obtain valuable insights on social stereotypes associated with each attribute.
Moreover, we can compute and balance the regard distributions to improve the fairness of the dataset, e.g.\ ensuring negative or offensive descriptions are not overly represented in the data.

 While we acknowledge several difficulties and challenges still remain (as we discuss in the following section), we hope this work will be a fundamental step towards understanding and addressing inherent ethical risks of LLM pretraining.

\section*{Limitations}
\label{sec:limitations}

While our taxonomy of protected attributes covers a reasonably wide range of demographics, it does not currently support attribute classes such as \textit{age}, \textit{marital status}, and \textit{political belief}, nor our list of attributes in each class is comprehensive, e.g.\ many attributes are still missing in the class of \textit{nationality}, \textit{disability}, etc.
Expanding and refining the taxonomy is a subject of future study, which we expect can be approached by leveraging LLMs and human-curated resources \citep{smith2022m}.

Regarding WSD, we found that Gloss-RoBERTa can make accurate predictions in many cases but still struggle in disambiguating several attributes.
In fact, there may also be genuine ambiguity which are difficult to be resolved even by humans.
Establishing clearer criteria and improving the robustness of WSD also remain as open problems.

In terms of regard classification, our current classifier is distilled from (i.e.\ relies on the judgements of) Mixtral-8x7B-Instruct, which themselves may be biased in some undesirable ways.
While we expect this issue can be alleviated by collecting multiple judgements from a diverse pool of human annotators or LLMs, developing a more reliable regard classier remains an important future work.

As for bias mitigation, we obtained promising results on reducing negative word associations through regard balancing: however, we've not conducted a full ablation study of pretraining LLMs from scratch on the regard balanced datasets.
Due to the extreme cost of the above experiments,  we leave it as an opportunity for future study.

Finally, due to the limitation of space, we provide an in-depth discussion on related work in Appendix \ref{sec:related_work}.
While bias analysis and mitigation is a widely studied topic in NLP and AI safety, we believe our work proposes a unique and promising approach in addressing these problems in the challenging area of LLM pretraining.

\bibliography{custom}

\appendix

\begin{table*}[t]
\centering
\begin{adjustbox}{max width=0.95\textwidth}
\setlength\tabcolsep{8pt}
\begin{tabular}{ccc}
\hline
Attribute Class & Attribute Keywords & Example Gloss (\quotes{a person ...}) \\
\hline
\hline
dietary habits & vegan, vegetarian & who is a vegan \\
\hline
disability & autistic, blind, deaf, depression, disabled, wheelchair & who is autistic \\
\hline
economic status & poor, rich & who is economically poor \\
\hline
fertility status & fertile, infertile & who is able to conceive children \\
\hline
gender/sexuality & female, male, nonbinary, queer, trans & of female gender \\
\hline
\multirow{3}{*}{nationality} & afghan, argentine, armenian, australian, austrian, belgian, & \multirow{3}{*}{of Afghan nationality}  \\
& brazilian, bulgarian, canadian, chilean, chinese, colombian,  & \\
& croatian, cuban, danish, dominican, egyptian, ... (46 more) & \\
\hline
physical traits & overweight, underweight & who is overweight \\
\hline
race/ethnicity & african, arab, asian, black, hispanic, latino, white & of African race/ethnicity \\
\hline
religion & buddhist, christian, hindu, jewish, muslim & who believes in Buddhism \\
\hline
residence & rural, suburban, urban & who lives in rural area \\
\hline
\end{tabular}
\end{adjustbox}
\caption{Our taxonomy of protected attributes. 
Each of the 97 attribute keywords corresponds to a protected attribute, which are categorized into 10 attribute classes.
Each gloss used for word sense disambiguation (\cref{subsec:protected_attribute_detection}) is crafted as a continuation of the template phrase: \quotes{\textit{a person ...}}.}
\label{tab:taxonomy}
\end{table*}

\section{Method Details}
\label{sec:method_details}

\subsection{Taxonomy of Protected Attributes}
\label{subsec:taxonomy}

In Table \ref{tab:taxonomy}, we provide an overview of our protected attribute taxonomy.
In this study, we collected a total of 97 keywords which are categorized into 10 attribute classes.
For the curation of the taxonomy, we referenced the diversity \& inclusion policies of multiple international companies to define attribute classes that must be protected from discrimination.
Then, we leveraged existing resources (e.g.\ Mixtral models) to expand the list of candidate keywords followed by manual curation.
The gloss for each keyword is generated based on hand-crafted templates followed by human verification.

\subsection{Mixtral Prompts}
\label{subsec:mixtral_prompts}

\begin{table*}[t]
\centering
\begin{adjustbox}{max width=0.9\textwidth}
\setlength\tabcolsep{8pt}
\begin{tabular}{p{1.08\textwidth}}
\hline
\textbf{WSD for Protected Attribute Detection (\cref{subsec:protected_attribute_detection})} \\
\hline
Below is a text including the keyword \quotes{\keyword}. Determine whether the keyword indicates a person (or people) \gloss. \\
- Focus on the keyword \quotes{\keyword}. Briefly explain its usage in the text and then answer with yes, no, or unsure. \\
- Answer yes if there is a clear reference to the person (or people) \gloss. \\
- Answer no if there is no such reference, e.g.\ the keyword describes *non-human* entities (e.g.\ animals, objects, events) or does not indicate the individual(s) \gloss. \\
- Answer unsure otherwise, e.g.\ if the reference is unclear, indirect, or implicit. \\
\\
Text: \passage \\
\\
After examining the text: \\
- Briefly justify your answer up to 100 words, and then answer with yes, no, or unsure. \\
- Conclude your response in the format: \quotes{Therefore, the answer is ...}. \\
\hline
\hline
\textbf{Regard Classification (\cref{subsec:regard_classification})} \\
\hline
What best describes the regard towards the person (or people) \gloss \:in the following text? Focus on the individual(s) indicated by the keyword \quotes{\keyword} and answer from the following options:\\
- Positive: The regard for the individual(s) in the text is predominantly positive, e.g.\ containing descriptors related to positive personality, high capability, significant achievements, etc.\\
- Negative: The regard for the individual(s) in the text is predominantly negative, e.g.\ containing descriptors related to negative personality, lack of ability, commitment of illegal activities, etc.\\
- Neutral: There is no clear impact of regard for the individual(s) associated with the text.\\
\\
Text: \passage \\
\\
After examining the text: \\
- Start your response with the answer (positive, negative, or neutral), followed by justification up to 100 words. \\
\hline

\end{tabular}
\end{adjustbox}
\caption{Mixtral prompts used in word sense disambiguation (WSD) for protected attribute detection (\cref{subsec:protected_attribute_detection}) and regard classification (\cref{subsec:regard_classification}). Similar to the input format in Table \ref{tab:gloss_roberta}, \passage \, indicates the text containing the keyword, \keyword \, the target keyword, and \gloss \, the keyword's defined gloss.}
\label{tab:mixtral_prompts}
\end{table*}

\begin{table*}[t]
\centering
\begin{adjustbox}{max width=1.0\textwidth}
\begin{tabular}{lc}
\hline
\multicolumn{1}{c}{Example Sentence (Keyword in \textbf{Bold})} & Prediction \\
\hline
\hline
My wife is \textbf{vegan} so we went to this place and she really loved the food. & \multirow{6}{*}{\checkmark \,Protected Attribute} \\
I know that the advocates for the \textbf{blind} and visually impaired will continue his work, ... & \\
Our aim is to help the \textbf{poor} and marginalized communities in Ghana to have access to education ... & \\
She is thought to be the youngest certified \textbf{female} yoga instructor in the United States. & \\
To be alone in secret with the Father should become one of the highest joys for a \textbf{Christian}. & \\
But \textbf{rural} children in particular are more likely to be driven to school or other activities, ... & \\
\hline
This water-based, non-toxic, \textbf{vegan} nail color has been formulated especially for kids. & \multirow{6}{*}{$\times$ \,Non-Protected Attribute} \\
We're all \textbf{blind} to our own mistakes, and a fresh pair of eyes can do wonders for our manuscripts. & \\
It has received \textbf{poor} reviews from critics and viewers, who have given it an IMDb score of 6.1. & \\
Did you know that \textbf{female} fireflies can’t fly? & \\
\textbf{Christian} Byrt is an award winning photographer based in Bunbury. & \\
Road connectivity of \textbf{rural} areas to urban areas is yet to be accomplished. & \\
\hline
\end{tabular}
\end{adjustbox}
\caption{Illustrative examples of word sense disambiguation (WSD) for protected attribute detection (\cref{subsec:protected_attribute_detection}) where both Mixtral-8x22B and Gloss-RoBERTa agreed on the predictions.}
\label{tab:wsd_examples}
\end{table*}

\begin{table*}[t]
\centering
\begin{adjustbox}{max width=1.0\textwidth}
\begin{tabular}{lc}
\hline
\multicolumn{1}{c}{Example Sentence (Keyword in \textbf{bold})} & Prediction \\
\hline
\hline
Our \textbf{female} guide was calm, competent and interesting whilst our male guide was more hyperactive. & \multirow{3}{*}{Positive} \\
We become quickly amazed by the \textbf{Japanese} people manners, politeness, and way of life from the beginning ... & \\
Although she was \textbf{overweight}, he still loved every part of her as she learned how to manage her health. & \\
\hline
The girl is described as a white \textbf{female}, standing 5-foot-2 and weighing 120 pounds with blue eye ... & \multirow{3}{*}{Neutral} \\
Some cultures traditionally swam in their undergarments, and for the \textbf{Japanese}, that meant the fundoshi, ... & \\
Many \textbf{overweight} people lose weight by this one rule – especially if they have previously consumed a lot of ... & \\
\hline
Debora Green is an American physician and \textbf{female} Villain who pleaded no contest to setting a 1995 fire ... & \multirow{3}{*}{Negative} \\
This showed horrific scenes of cruelty and deception by the \textbf{Japanese} authorities who want to hide this from ... \\
... and the general description that people have of her is that she is \textbf{overweight} and unattractive. & \\
\hline
\end{tabular}
\end{adjustbox}
\caption{Illustrative examples of regard classification towards protected attributes (\cref{subsec:regard_classification}) where both Mixtral-8x7B and our RoBERTa-based classifier agreed on the predictions.}
\label{tab:regard_examples}
\end{table*}

In Table \ref{tab:mixtral_prompts}, we show the actual prompts used for data annotation based on Mixtral models.

Regarding WSD for protected attribute detection (\cref{subsec:protected_attribute_detection}), we applied chain-of-thought (CoT) prompting \citep{wei2022chain} to first explain the keyword usage in the context, which consistently improved the prediction quality.
Based on Mixtral's response, we considered the answer of \textit{yes} as protected attribute and \textit{unsure/no} as non-protected attribute.
In Table \ref{tab:wsd_examples}, we show several examples of WSD results where both Mixtral-8x22B and Gloss-RoBERTa agreed on the predictions.

As for regard classification (\cref{subsec:regard_classification}), we generally followed the criteria of \citet{sheng-etal-2019-woman} and explicitly included them in the prompt.
In Table \ref{tab:regard_examples}, we show several illustrative examples where both Mixtral-8x7B and our RoBERTa-based classifier agreed on the predictions.


\begin{table*}[t!]
\setlength\extrarowheight{-1.5pt}
\centering
\begin{adjustbox}{max width=1.0\textwidth}
\setlength\tabcolsep{3.5pt}
\begin{tabular}{
    cllll
}
\hline
\multirow{2}{*}[-0.7ex]{\shortstack[c]{Protected\\Attribute}} & \multicolumn{1}{c}{\multirow{2}{*}[-0.7ex]{\shortstack[c]{Frequency Bias\\(eq. (\ref{eq:bias_score}))}}} & \multicolumn{3}{c}{\raisebox{-0.2ex}{Frequency+Regard Bias (eq. (\ref{eq:regard_bias_score}))}} \\
\cmidrule(lr){3-5}
 & & \multicolumn{1}{c}{$r=$ Positive} & \multicolumn{1}{c}{$r=$ Negative} & \multicolumn{1}{c}{$r=$ Neutral} \\
\hline
\hline
\multirow{4}{*}{chinese} & {\small chinese, xi, beijing, china, kong,} & {\small scientists, jack, ma, artificial, sun,} & {\small sensitive, edited, selling, theft, do-} & {\small evaluate, examined, po-} \\
& {\small hong, ma, apple, investors, korea,} & {\small richest, optimism, innovation, \green{wise},} & {\small llars, sweeping, spreading, profit,} & {\small pulations, examine, sid-} \\
& {\small korean, consumers, epidemic, ou-} & {\small medicine, \green{ambitious}, technological,} & {\small amounts, restrictions, apple, debt,} & {\small elines, backgrounds, sa-} \\
& {\small tbreak, belt, russians, accepting, ..} & {\small boost, enhance, engineering, ...} & {\small canceled, sell, impose, racist, ...} & {\small mple, comparative, ...} \\
\hline
\multirow{4}{*}{iraqi} & {\small iraqi, iraq, bush, coalition, civil-} & {\small sacrifice, veteran, \green{brave}, determin-} & {\small executed, shooting, threw, abuse,} & {\small assess, mechanism, dis-} \\
& {\small lians, forces, civilian, fighters, d-} & {\small ation, minds, capable, volunteer, c-} & {\small prison, murder, throwing, prison-} & {\small cuss, accordance, decli-} \\
& {\small estruction, combat, abu, troops,} & {\small ourage, helping, liberation, stabili-} & {\small ers, fired, crimes, suicide, blame-} & {\small ned, request, discussed,} \\
& {\small weapons, syrian, refugees, ...} & {\small ty, democracy, optimistic, relief, ...} & {\small d, dead, firing, least, killed, ...} & {\small postponed, walks, ..} \\
\hline
\multirow{4}{*}{mexican} & {\small mexico, \green{slim}, texas, \red{drug}, fox,} & {\small guitar, painter, colors, iconic, vibra-} & {\small lord, \red{drug}, \red{racist}, \red{illegal}, gun, ag-} & {\small populations, examined,} \\
& {\small carlos, california, el, cuban, d-} & {\small nt, superstar, richest, y, shape, hom-} & {\small ents, drugs, guns, pounds, judge,} & {\small cuban, ancestry, regar-} \\
& {\small iego, los, y, angeles, las, lord,} & {\small age, painted, actress, honor, proud-} & {\small federal, customs, healine, deport-} & {\small dless, examine, origin,} \\
& {\small immigration, vegas, san, del, ..} & {\small ly, hollywood, starring, loves, ...} & {\small ation, agent, trump, wall, ...} & {\small agreement, identify, ...} \\
\hline
\multirow{4}{*}{pakistani} & {\small pakistan, khan, afghanistan, fi-} & {\small pioneer, designer, credited, culinary,} & {\small firing, fake, violation, alleged, at-} & {\small criteria, apply, sideline-} \\
& {\small ring, delhi, terror, bin, india, b-} & {\small nobel, education, entertainment, ox-} & {\small tacked, \red{terrorist}, blamed, banned, } & {\small s, applications, submit,} \\
& {\small angladesh, frontier, cia, muslim,} & {\small ford, survived, drama, teenager, \green{bra-}} & {\small district, killed, lone, arrested, pos-} & {\small eligible, counterpart, d-} \\
& {\small dawn, posts, northwest, raid, ...} & {\small \green{ve}, actress, advocate, commentator, ..} & {\small ts, targeted, allegedly, ...} & {\small elegation, scholarship, ..} \\
\hline
\multirow{4}{*}{ukrainian} & {\small ukrainian, ukraine, inquiry, joe,} & {\small \green{intelligent}, courage, boxing, faithful,} & {\small conspiracy, investigative, corrupt-} & {\small criteria, submit, discus-} \\
& {\small investigate, complaint, trump,} & {\small beautiful, tender, dignity, fights, par-} & {\small ion, lawmaker, imprisonment, al-} & {\small sed, accordance, teleph-} \\
& {\small phone, russia, bride, investiga-} & {\small ticipant, software, honored, lady, re-} & {\small legations, violations, positions, p-} & {\small one, presidents, depen-} \\
& {\small tions, call, russians, ...} & {\small liable, sciences, resilience, ...} & {\small olitically, prosecutor, scrutiny, ...} & {\small ds, procedures, ...} \\
\hline
\end{tabular}
\end{adjustbox}
\caption{Results of our bias analyses for class $A=$ \textit{nationality}. Words $w \in V$ are sorted in descending order based on the frequency bias score (eq. (\ref{eq:bias_score})) and frequency+regard bias score (eq. (\ref{eq:regard_bias_score})). Positive stereotypes listed in SeeGULL is highlighted in \green{green} and negative stereotypes in \red{red}.}
\label{tab:bias_analysis_nationality}
\end{table*}

\begin{table*}[t!]
\centering
\begin{adjustbox}{max width=0.75\textwidth}
\setlength\tabcolsep{9pt}
\begin{tabular}{
    cccc
}
\hline \\[-12pt]
 & Recall@ & Frequency Bias & Frequency+Regard Bias \\
\hline
\hline
\multirow{5}{*}{\shortstack[c]{Positive Stereotypes\\(\textit{mean offensiveness score} $= -1$)}} & 50 & 2.67 & \textbf{5.01} \\
 & 100 & 4.00 & \textbf{7.79} \\
 & 200 & 7.23 & \textbf{13.24} \\
 & 300 & 9.79 & \textbf{16.80} \\
 & 500 & 15.80 & \textbf{24.69} \\
 \hline
 \multirow{5}{*}{\shortstack[c]{Negative Stereotypes\\(\textit{mean offensiveness score} $\geq 1$)}} & 50 & 3.68 & \textbf{13.97} \\
 & 100 & 9.56 & \textbf{17.65} \\
 & 200 & 13.97 & \textbf{33.09} \\
 & 300 & 16.18 & \textbf{38.97} \\
 & 500 & 25.00 & \textbf{47.06} \\
\hline
\end{tabular}
\end{adjustbox}
\caption{Results of the alignment with SeeGULL stereotypes. For all thresholds of recall@k, our frequency+regard bias analysis consistently outperforms the baseline based on the frequency bias analysis.}
\label{tab:seegull_alignment}
\end{table*}

\section{Bias Analysis Details}
\label{sec:bias_analysis_details}

To conduct our bias analysis, we must first define the set of vocabulary $V$.
For the purpose of identifying social stereotypes, we focused on common words which are likely to appear with any protected attribute.
Specifically, we first computed 20K most frequent words $V_a$ which appear with each attribute $a \in A$. 
Then, we took the intersection among them as the final set, i.e.\ $V = \bigcap_{a \in A} V_a$.\footnote{Note that we focus on comparing attributes in the same class $A$ and avoid comparing among different classes (e.g.\ \textit{female} and \textit{blind} people) to ensure that attributes are measurable by the same standard \citep{blodgett-etal-2021-stereotyping}.}

In Table \ref{tab:bias_analysis_nationality}, we show (partial) results of our bias analysis (\cref{subsec:bias_analysis}) for the class $A =$ \textit{nationality}.
Due to the large number of attributes in this class, we restricted our analysis to 53 nationalities which are covered in the SeeGULL dataset \citep{jha-etal-2023-seegull}.\footnote{\url{https://github.com/google-research-datasets/seegull}.}
SeeGULL is a high quality and broad coverage resource containing prevalent stereotypes, which we use for the verification of our analysis.

Specifically, we utilize their \textit{offensiveness score} annotations and consider the stereotype in their list as positive if the mean score is $-$1 (e.g.\ \quotes{\textit{wise}} for \textit{chinese}) and negative if it is greater than or equal to 1 (e.g.\ \quotes{\textit{greedy}} for \textit{japanese}). 
Considering these stereotypes as the ground truths, we measured the recall@k for the 53 nationalities based on the vocabulary sorted by the frequency bias and frequency+regard bias scores.\footnote{As for the frequency+regard bias scores, we set the regard $r$ = positive/negative in eq. (\ref{eq:regard_bias_score}) to compute the alignment with SeeGULL's positive/negative stereotypes, respectively.}

In Table \ref{tab:seegull_alignment}, we show the results of the alignment with SeeGULL stereotypes.
From these results, we can verify that frequency+regard bias scores can identify positive/negative stereotypes more accurately, as (partially) illustrated in Table \ref{tab:bias_analysis_nationality}.
In fact, our analysis can detect potential stereotypes that may be even missing in SeeGULL (e.g.\ \quotes{\textit{technological}} or \quotes{\textit{racist}} against \textit{chinese} people), demonstrating the effectiveness of our approach. 
\section{Related Work}
\label{sec:related_work}

Regarding the safety/fairness of LLM pretraining datasets, \textit{toxicity} \citep{gehman-etal-2020-realtoxicityprompts} is a related yet distinct topic from social biases.
For instance, toxic contents include expressions such as \quotes{\textit{White people are racists}}, which are directly harmful by themselves and can be removed at the instance-level based on toxicity classifiers \citep{longpre-etal-2024-pretrainers}.
In contrast, the problem of social biases include expressions which may be harmless by themselves but would be collectively harmful if overrepresented in the data.
For instance, expressions such as \quotes{\textit{They were protesting racist white police officers ...}} are observed in high-quality texts (e.g.\ news articles) and seem harmless by themselves, but they may promote undesirable stereotypes if excessively represented in the corpora.
Unlike toxic contents, such expressions are difficult to be removed at the instance-level and are more desirable to be properly balanced, e.g.\ through downsampling.

For relatively small-scale datasets (e.g.\ for finetuning), existing works propose data augmentation methods to enhance the fairness among protected attributes \citep{lu2020gender,qian-etal-2022-perturbation}.
However, such approaches are often difficult to be applied on massive-scale datasets, and developing scalable and effective methods for bias analysis and mitigation remain as open challenges.

Finally, biased generations from LLMs can be mitigated at the post-training step \citep{schick2021self,ganguli2023capacity} or through external guardrails \citep{inan2023llama,dong2024safeguarding}.
While these approaches can substantially suppress the exhibition of biased outputs, it remains unclear whether undesirable biases are effectively removed or merely hidden by the models \citep{gonen-goldberg-2019-lipstick}.
Therefore, we believe understanding and alleviating social biases in the pretraining data remain as crucial problems, both from academic and industrial perspectives \citep{dodge2021documenting,luccioni-viviano-2021-whats,feng-etal-2023-pretraining,esiobu-etal-2023-robbie,baack2024critical}.

\end{document}